\documentclass[a4paper,twocolumn,10pt]{article}
\usepackage{eurosis}
\usepackage{url}
\usepackage{natbib}

\usepackage{subfigure}
\usepackage{graphicx}
\usepackage[ruled,vlined,linesnumbered]{algorithm2e}
\usepackage{url}
\usepackage{algorithmic}
\usepackage{amsfonts}
\usepackage{amsmath}
\usepackage{cite}
\usepackage{array}

\usepackage{caption}
\usepackage{float}
\usepackage{stfloats}
\usepackage{comment}
\usepackage{multirow}

\makeatletter
\def\NMSB@titlestyle{simple}
\def\NMSB@stylesimple#1#2{\if@nobreak\else\bigskip\fi\textbf{\large #1#2}}
\makeatother

\bibpunct[; ]{(}{)}{,}{a}{}{;}

\begin{document}

\title{ON THE EFFECTS OF THE VARIATIONS IN NETWORK CHARACTERISTICS IN CYBER PHYSICAL SYSTEMS}


\author{
G{\'e}za Szab{\'o}, S{\'a}ndor R{\'a}cz\\
Ericsson Research, Budapest, Hungary\\
Email: \{geza.szabo,sandor.racz\}@ericsson.com
\and
J{\'o}zsef Pet\H{o}\\
Budapest University of Technology and Economics, Hungary\\
Email: pjoejoejoe@gmail.com
\and
Rafael Roque Aschoff\\
Pernambuco Federal Institute of Education, Science, and Technology, Brazil\\
Email: rafael.aschoff@gprt.ufpe.br
}

\date{}

\maketitle

\thispagestyle{empty}


\begin{abstract}
The popular robotic simulator, Gazebo, lacks the feature of simulating the effects of control latency that would make it a fully-fledged cyber-physical system (CPS) simulator.
The CPS that we address to measure is a robotic arm (UR5) controlled remotely with velocity commands. The main goal is to measure Quality of Control (QoC) related KPIs during various network conditions in a simulated environment. 
We propose a Gazebo plugin \citep{latencyplugingithub} to make the above measurement feasible by making Gazebo capable to delay internal control and status messages and also to interface with external network simulators to derive even more advanced network effects.
Our preliminary evaluation shows that there is certainly an effect on the behavior of the robotic arm with the introduced network latency in line with our expectations, but a more detailed further study is needed.
\end{abstract}

\section{INTRODUCTION}\label{intro}

A cyber-physical system~(CPS) is a mechanism controlled or monitored by computer-based algorithms, tightly integrated with the internet and its users. 
Unlike more traditional embedded systems, a full-fledged CPS is typically designed as a network of interacting elements with physical input and output instead of as standalone devices.
For tasks that require more resources than are locally available, one common mechanism is that nodes utilize the network connectivity to link the sensor or actuator part of the CPS with either a server or a cloud environment, enabling complex processing tasks that are impossible under local resource constraints.
Among the wide diversity of tasks that CPS is applied we focus on robot control in this paper.

Currently, one of the main focus of cloud based robotics is to speed up the processing of input data collected from many sensors with big data computation. Another approach is to collect various knowledge bases in centralized locations e.g.,~possible grasping poses of various 3D objects. 

Another aspect of cloud robotics is the way in which the robot control related functionality is moved into the cloud. The simplest way is to run the original robot specific task in a cloud without significant changes in it. For example, in a Virtual Machine~(VM), in a container, or in a virtualized Programmable Logic Controller~(PLC). Another way is to update, modify or rewrite the code of robot related tasks to utilize existing services or APIs of the cloud. The third way is to extend the cloud platform itself with new features that make robot control more efficient. These new robot-aware cloud features can be explicitly used by robot related tasks (i.e.~new robot-aware services or APIs offered by cloud) or can be transparent solutions (e.g.,~improving the service provided by the cloud to meet the requirement of the robot control).

Designing cyber-physical systems is challenging because of a) the vast network and information technology environment connected with physical elements involves multiple domains such as controls, communication, analog and digital physics, and logic and b) the interaction with the physical world varies widely based on time and situation. 
To ease the design of CPS, robot simulators have been used by robotics experts. A well-designed simulator makes it possible to rapidly test algorithms, design robots, perform regression testing, and train AI system using realistic scenarios. 

There are various alternatives, sets of tools that make it possible to put together a CPS simulation environment, but it is very difficult, needs a lot of interfacing with various tools and impractical. 
The requirements of a widely applicable CPS are the following:
\begin{itemize}
 \item{Should be modular in terms of interfacing with the CPS}
 \item{Should be modular in terms of interfacing with network simulator, realization environment}
 \item{Should be able to cooperate with widely applied environments}
\end{itemize}

We chose Gazebo as our target robot simulation environment that we intend to extend with new functionalities to make it capable of being applied as a CPS. 
Gazebo~\citep{gazebourl} offers the ability to accurately and efficiently simulate populations of robots in complex indoor and outdoor environments. It has a robust physics engine, high-quality graphics, and convenient programmatic and graphical interfaces. Gazebo is free and widely used among robotic experts.

The main challenge with the design principle of Gazebo is that the control of actuators is deployed and run practically locally to the actuators. In this case, there is no need to consider the effects of a non-ideal link between the actuator and the controller. Considering the CPS context, as controllers are moved away from actuators, it becomes natural and even necessary to analyze the effects of the network link between them.

Gazebo has a plugin system that we target to use to provide us an interface to our modular network simulation environment.
The goal of this paper is to show the design principles of the network plugin and provide the research community with a tool for further research in CPS.


\section{THE MEASUREMENT SETUP THAT WE GO FOR}
The CPS that we address to measure is a robotic arm (UR5~\citep{ur5}) controlled remotely with velocity commands. The main goal is to measure Quality of Control (QoC) e.g.,~cumulated PID error during trajectory execution, cumulated difference in joint space between the executed and calculated trajectories, etc. related KPIs during various network conditions in this setup. 

\begin{figure}
\centering
\includegraphics[width=.45\textwidth]{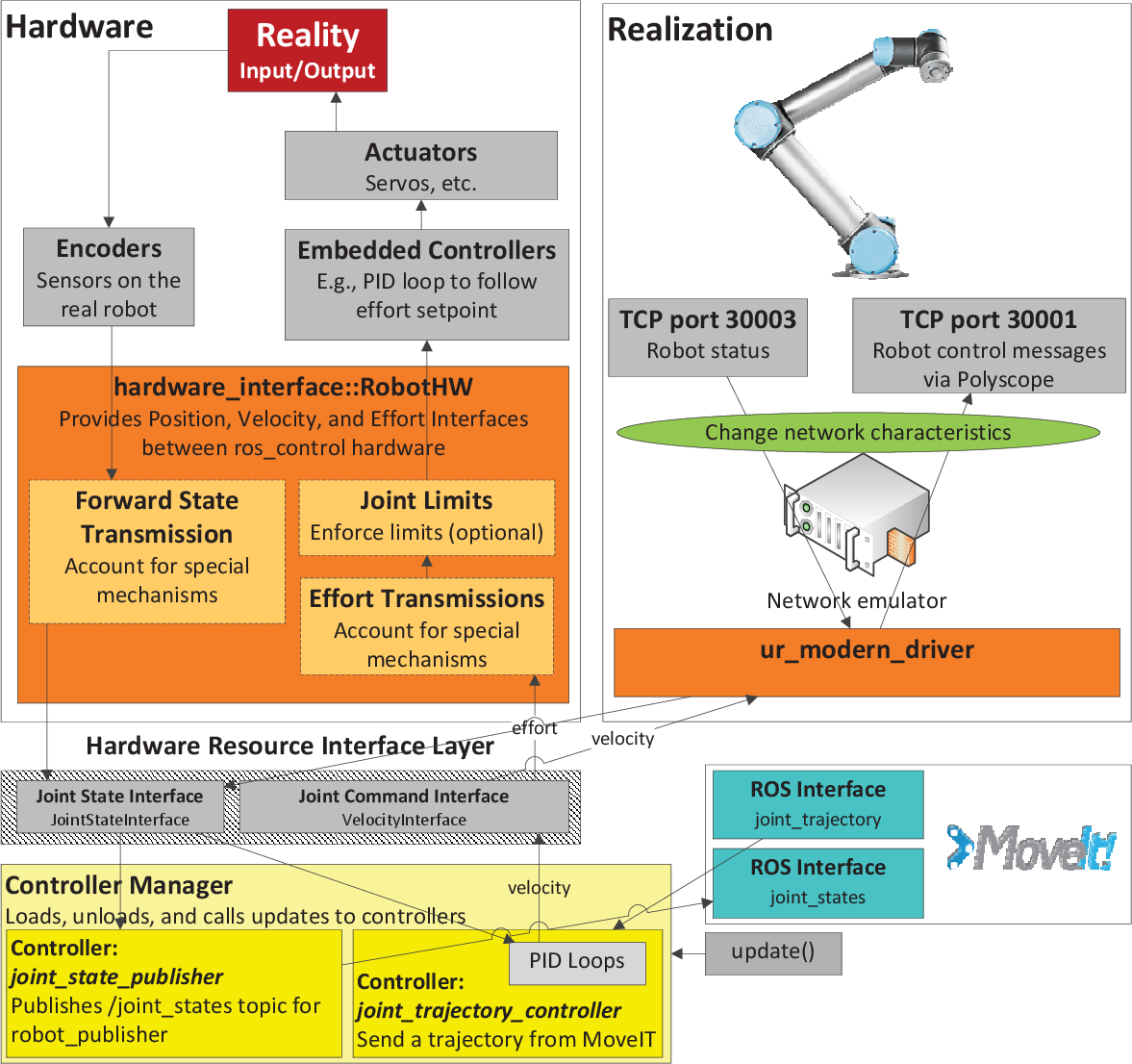}
\caption{Target architecture to be realized with simulator}
\label{hw}
\end{figure}

Figure~\ref{hw} shows the use case with real hardware that we target to simulate in Gazebo. The left side of the figure (Hardware) shows the same data elements described in \citep{roscontrol}, whereas the right side of the picture (Realization) uses the same colors for the boxes to describe a specific realization.
In the specific case, the UR5 can be accessed via TCP/IP ports 50001 to send command messages and port 50003 to read the robot status messages. The trajectories are computed by MoveIt~\citep{MoveIt}. MoveIt sends trajectories to the controller manager which starts a velocity controller (yellow), a specific type of \texttt{ros\_control}. The \texttt{ur\_modern\_driver}~\citep{andersen2015optimizing} implements the hardware resource interface layer by simply copying the velocity control packets to the proper TCP sockets. A middle node can be deployed between the robot driver and the robot (green) that can alter the network characteristics. 

A trivia approach to setup the above architecture in a simulation environment is provided by Universal Robots. Universal Robots simulator software \citep{URSim} is a java software package that makes it possible to create and run programs on a simulated robot, with some limitations. 
The limitation of this solution is that it is capable to simulate only one robot. There is no chance to integrate the robot in complex environments as you can configure with Gazebo e.g.,~interacting with other mechanical elements in the workspace, check collisions with the environment, etc.

\begin{figure*}
\centering
\includegraphics[width=.99\textwidth]{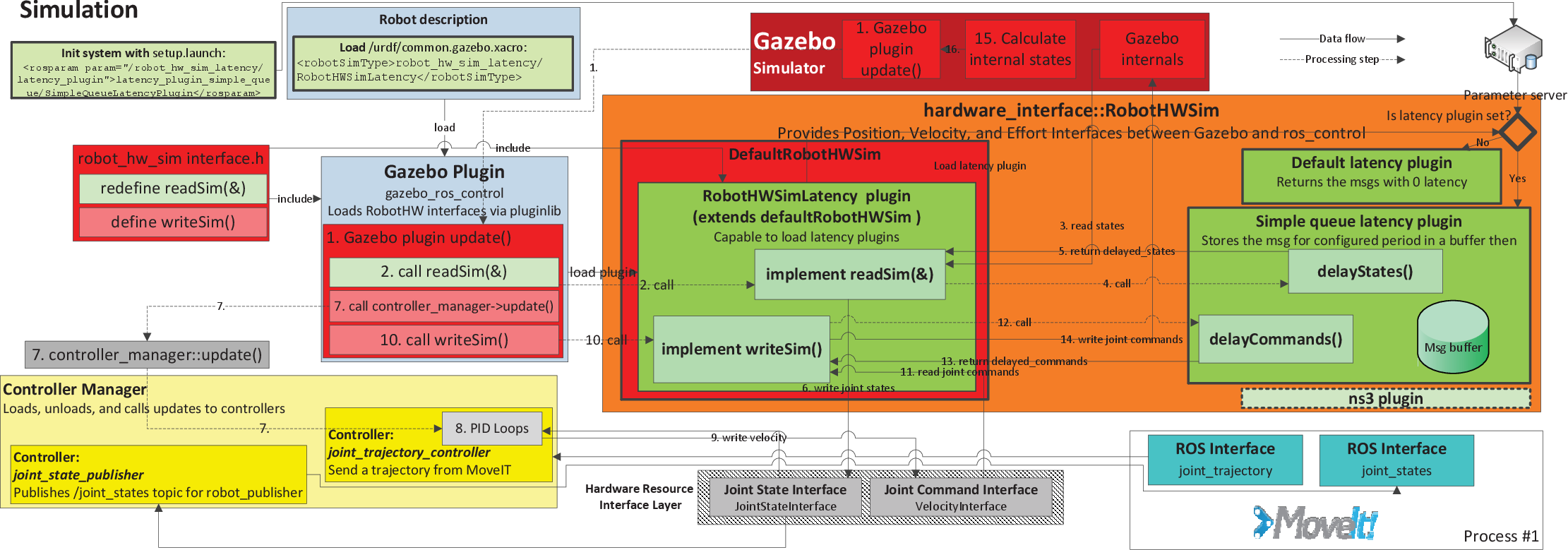}
\caption{Gazebo architecture}
\label{gazebo}
\end{figure*}

\section{MOTIVATION AND RELATED WORK}\label{sec:related_work}
\subsection{COMPETITIONS}
A frontier method to push research groups to their limits is to organize competitions.
DARPA, a research group in the U.S. Department of Defense, announced the DARPA Robotics Challenge with a US \$2 million dollar prize for the team that could produce a first responder robot performing a set of tasks required in an emergency situation.
During the DARPA Trials of December 2013, a restrictive device was inserted into the control computers of each competing team and the computer that formed the 'brain' of the robot. 
The intent of the network degradation was to roughly simulate the kind of less than perfect communications that might exist during those kinds of emergency or disaster situations in which these robots would be deployed.
The restrictive device --, a Mini Maxwell network emulator from InterWorking Labs -- alternated between a 'good' mode and a 'bad' mode of network communication, every sixty seconds. 'Good' minutes permitted communications at a rate of 1~Mbps (in either direction) and a base delay of 50~ms (in each direction.) 'Bad' minutes permitted communications at a rate of 100~Kbps (in either direction) and a base delay of 500~ms (in each direction.) 
At the end of each minute, a transition occurred from bad-to-good or good-to-bad. A side effect of these transitions was packet-reordering. 

The impact of network degradation on the teams was larger than expected.
Informal feedback suggested that several teams did not realize that rate limitation induces network congestion or the ramifications of that congestion. 
Some teams appeared to have been surprised by the behavior of the network protocol stacks, particularly TCP stacks, in the operating systems underneath their code. \citep{darpa}
The above experiences would have been probably less striking to the teams if they were able to test the network characteristics changes in a simulation environment.

A recent competition Agile Robotics for Industrial Automation Competition (ARIAC)\citep{ariac} targets industrial related applications.
ARIAC is a simulation-based competition is designed to promote agility in industrial robot systems by utilizing the latest advances in artificial intelligence and robot planning. 
There is no tricky network environment in the ARIAC competition. The industry relies on robust low-delay protocols. That is why it is an interesting aspect to see what happens when those links and protocols are exchanged. For instance, what are the possible performance improvements or degradation when the control or sensors data processing in an industrial scenario are moved further away from the actuators and how different protocols would fare under various network characteristics?

\subsection{WHY GAZEBO?}
In both of the above competitions, Gazebo provided the simulation infrastructure. 
In a more structured study about the level of how wide-spread the various simulator tools were done in \citep{7041462}. It showed that Gazebo emerges as the best choice among the open-source projects.



Authors of \citep{omnetros} describes some early experiments in linking the OMNET++ simulation framework with the ROS middleware for interacting with robot simulators in order to get within the OMNET++ simulation a robot's position which is accurately simulated by an external simulator based on ROS. The motivation is to use well-tested and realistic robot simulators for handling all the robot navigation tasks (obstacle avoidance, navigation towards goals, velocity, etc.) and to only get the robot's position in OMNET++ for interacting with the deployed sensors. 
Our goal is the other way around, thus to introduce the effects of the network simulator into the robot simulator.

The roadmap of Gazebo development shows that version 9.0 arriving at 2018-01-25 will have support to integrate network simulation (ns-3 or EMANE). Further information regarding if this feature will be like \citep{omnetros} or the one we propose in this paper is not available yet.

\section{PROPOSED METHOD TO SIMULATE THE EFFECTS OF NETWORK CHARACTERISTICS}\label{section_proposed}

In ROS, topics~\citep{rostopics} are named buses over which nodes exchange messages. 
ROS currently supports TCP/IP-based and UDP-based message transport. 
ROS nodes are standalone executables running with individual process IDs in the operating system. 
One practical way to introduce latency in current ROS deployment is via defining network namespaces among nodes. For a certain namespace, custom delay, jitter, drop characteristics can be defined with \texttt{tc} like in \citep{nwnstc}.
The main issue is that there is a MoveIt node as an individual process, but the whole joint controller-actuator control loop is realized within Gazebo as one other process. The only topic based communication happens between the MoveIt and the monolith Gazebo process. So this kind of solution cannot be applied to our problem.

We have to dig deeper in the architecture of Gazebo and realize the CPS system within. To keep the architecture modular, we decided to implement the proposed method as a Gazebo plugin. While the setup most of these plugins fits well in the current Gazebo architecture and can be done via configuration files, there are still patches needed to be applied on core functional elements of the Gazebo code.

Figure~\ref{gazebo} shows the architecture of the proposed method. The coloring of the figure follows the way in \citep{roscontrol}. Green represents new added plugins, modules, functionalities. 
The working of the system is the following. 
As a first step --, a launch file that triggers the whole simulation to run -- setups a parameter on the ROS parameter server. This parameter defines the specific latency plugin that will be loaded.

The launch file initiates the Gazebo simulation. Gazebo loads the \texttt{gazebo\_ros\_control} plugin (left most blue box) that main purpose is to interface with the ROS controller manager. This module needed a small tweak. The original code passed the address of the messages from the controller manager to the simulation, performed the actions (read status, calculate commands) triggered by the \texttt{update()} function in a sequential manner. There was no modification of these input variables during the calculations in the original code. In our system, the messages are copied and stored to make it possible to perform further actions on the messages.

Gazebo loads configuration files from the \texttt{common.gazebo.xacro} file in which it is specified that our custom \texttt{RobotHWSimLatency} plugin should be loaded instead of the \texttt{DefaultRobotHWSim} plugin. Our \texttt{RobotHWSimLatency} plugin is the extension of the \texttt{DefaultRobotHWSim} plugin with modified read and write functions and with the task to load a custom latency plugin. The latency plugin to be loaded is the one that was setup by the parameter server. 
The current options include a) the default latency plugin that practically returns the messages with no introduced latency and b) the simple queue latency plugin. This later has a configurable size of the queue to store the messages in them. In each simulation tick (100Hz), the messages are shifted one position forward in the queue and when they reach the end of the queue they are provided to Gazebo as the currently valid message.
In the same way, an interface plugin to cooperate with external network simulators like ns3~\citep{ns3} can be also implemented here.
We described the detailed call sequence of the plugin system in details on \citep{latencyplugingithub}.

\begin{figure}
\centering
\includegraphics[width=.45\textwidth]{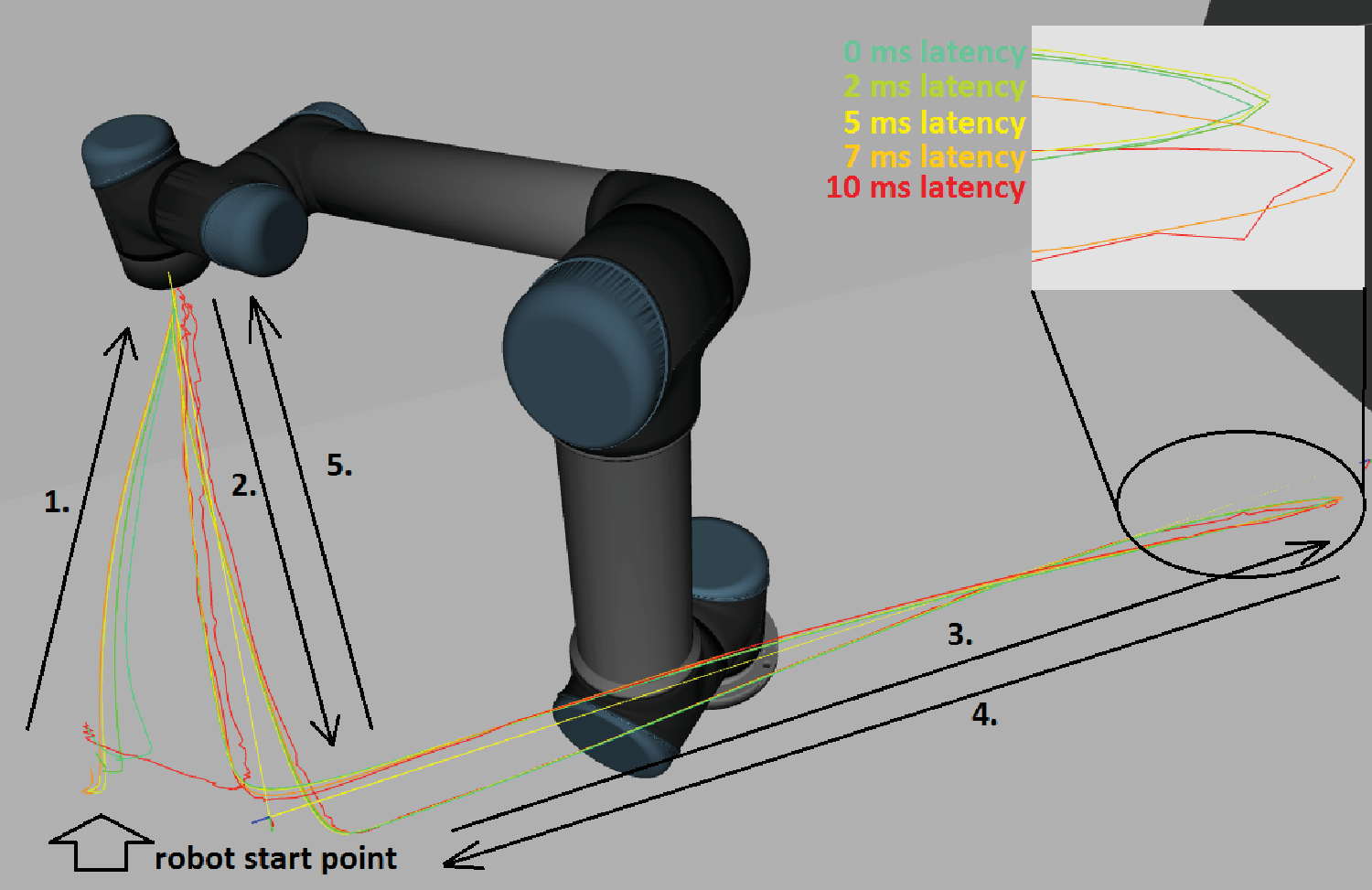}
\caption{The visualized trajectories}
\label{trajectories}
\end{figure}

\section{EVALUATION} \label{sec:evaluation}
We evaluated our proposed method on various Key Performance Indicators (KPIs).
The most straightforward evaluation is the visual inspection of the robotic arm movement. For this purpose, we loaded the robot model into rviz and used a ros package to visualize the markers on the way the robotic arm passed through. Figure~\ref{trajectories} is a screenshot from rviz which shows the visualized trajectories. The bottom left corner of the picture is the starting point of the robotic arm. It passes through the waypoints one-by-one from number 1 to 5. The black lines are the trajectories, while the lines with various colors show the effect of introducing latency into the system. The cyan color shows the reference scenario with 0 latency. In all other cases, we introduced latency in the system in both the command writing and status reading direction and rerun the trajectory planning and execution scenario.
The upper right corner of the picture shows a magnified part around the trajectories. The trajectories were planned with the \texttt{RRTConnectkConfigDefault} planner.

The visualized trajectories show the expected behavior of the system. Increasing the latency increases the deviance from the original trajectories. 
It should be noted that the planned trajectories are straight in Cartesian-space. To move along these trajectories the robotic arm needs complex movements in the joint-space, thus even the movement in a straight line causes deviation from the reference trajectory.
In the other way around, if the planned trajectories were straight in the joint-space, we would see a movement in circles by the robotic arm, but the effect of the latency was more negligible.

\begin{figure}
\centering
\includegraphics[width=.45\textwidth]{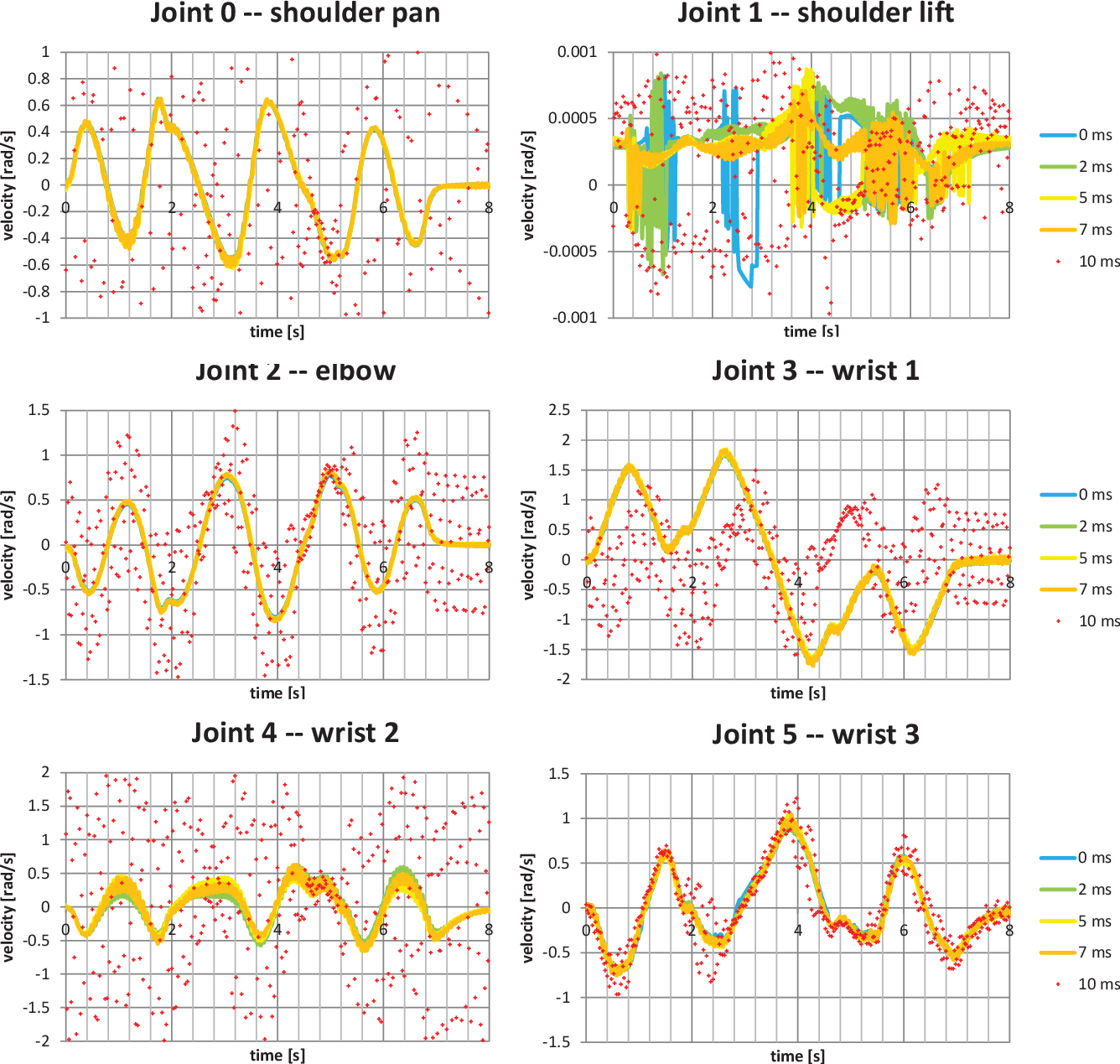}
\caption{The velocity commands sent to the robot}
\label{evaljoints}
\end{figure}

Figure~\ref{evaljoints} shows the velocity commands sent to the robot in the function of time. Analyzing the velocity commands in such details reveals that comparing the different scenarios are not straightforward for several reasons. One is that the planning is non-deterministic, and a slight difference during the initialization of the gazebo environment ends up with some different planned trajectory. The execution of the trajectories depends on the environment status as well, and it is never the same.
Joint~4 shows the expected effect on the velocity commands levels as well, thus the induced latency causes increased velocity command deviation compared to the reference scenario.
It is also a clear observation that around 10~ms latency, the system starts to get unstable. This is likely due to the various updating frequency parameters that Gazebo employs to run the simulation. It needs definitely further work to make it clear how the introduced latency affects other characteristics or behaviors, such as the robot commanding frequency, whole physical simulation steps, internal message timings.

\begin{figure}
\centering
\includegraphics[width=.35\textwidth]{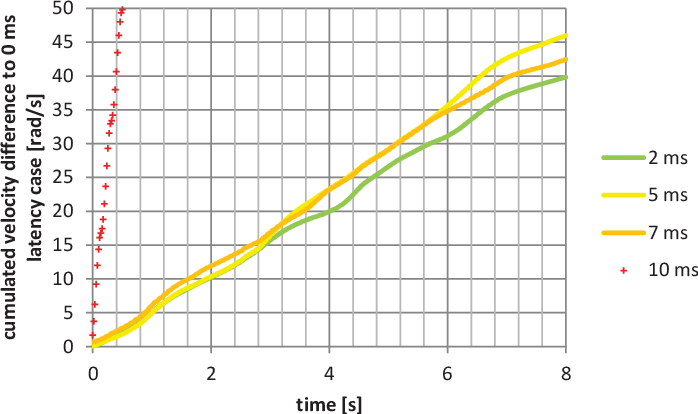}
\caption{The cumulated difference of the velocity commands comparing to the reference scenario}
\label{cumulatederror}
\end{figure}

Figure~\ref{cumulatederror} shows the cumulated difference of the velocity commands comparing to the reference scenario. The 2~ms latency scenario is the closest to the reference as it is expected. In the first 3~sec of the trajectory execution the 5~ms scenario is closer to the reference than the 7~ms scenario, but around 6~sec, the 5~ms scenario collects so much error that shows bigger deviation than the 7~sec scenario. The 10~sec scenario has another magnitude of error, and thus cut off the diagram after the first second.

\section{CONCLUSION AND FURTHER WORK}\label{sec:conclusion}
In this paper, we proposed a plugin \citep{latencyplugingithub} to extend the capabilities of the current Gazebo robotic simulator and turn it into a CPS system. The realization of the proposed method is a plugin to Gazebo. The plugin fits into the modular design of Gazebo. As of the interface is available, it eases to test various network effects on the robot control.
Based on our preliminary evaluations it does affect the QoC KPIs of the robot control.

The evaluation showed behavior which is expected and reasonable, but also cases which show that the whole system needs fine-tuning. We plan to evaluate the working mechanism of the system with the help of the ROS, gazebo and research communities.
We plan to do more extensive measurements with the tool. We plan to interface it with various radio network simulators and see the effects of the radio on the QoC KPIs.
In a similar way, we plan to investigate the how the system behaves when taking into account not only the network links characteristics but also the protocols for message exchanging.
We also plan to compare the level of similarity of the simulation to real robot HW controlled in a real radio network.
We are taking part in the ARIAC competition and we plan to evaluate if the tool can provide any advantage for us in any of the use cases of the competition.

\bibliographystyle{eurosis}
\bibliography{demo}


\end{document}